\documentclass[letterpaper, 10 pt, conference]{ieeeconf}  % Comment this line out if you need a4paper

\IEEEoverridecommandlockouts                              % This command is only needed if 
\usepackage{cite}
\usepackage{amsmath,amssymb,amsfonts}
\usepackage{algorithm} 
\usepackage{algpseudocode} 
\usepackage{graphicx}
\usepackage{multirow}
\usepackage{booktabs}
\usepackage{verbatim}
\usepackage{textcomp}
\usepackage{xcolor}

\title{\LARGE \bf
A post-selection algorithm for improving dynamic ensemble selection methods
}

\author{Paulo R.G. Cordeiro$^{1}$, George D.C. Cavalcanti$^{2}$ and Rafael M.O. Cruz$^{3}$ % <-this % stops a space
\thanks{$^{1}$Paulo R.G. Cordeiro is with Centro de Informática, Universidade Federal de Pernambuco, Recife, Brazil {\tt\small (prgc@cin.ufpe.br)}}%
\thanks{$^{2}$George D.C. Cavalcanti is with Centro de Informática, Universidade Federal de Pernambuco, Recife, Brazil {\tt\small (gdcc@cin.ufpe.br)}}
\thanks{$^{3}$Rafael M.O. Cruz is with Département de génie Logiciel et des TI, École de Technologie Supérieure, Montreal, Canada {\tt\small (rafael.menelau-cruz@etsmtl.ca)}}%
\thanks{$^{4}$https://github.com/prgc/ps-des}
}

\begin{document}

\maketitle
\thispagestyle{empty}
\pagestyle{empty}

%%%%%%%%%%%%%%%%%%%%%%%%%%%%%%%%%%%%%%%%%%%%%%%%%%%%%%%%%%%%%%%%%%%%%%%%%%%%%%%%
\begin{abstract}

Dynamic Ensemble Selection (DES) is a Multiple Classifier Systems (MCS) approach that aims to select an ensemble for each query sample during the selection phase. Even with the proposal of several DES approaches, no particular DES technique is the best choice for different problems. Thus, we hypothesize that selecting the best DES approach per query instance can lead to better accuracy. To evaluate this idea, we introduce the Post-Selection Dynamic Ensemble Selection (PS-DES) approach, a post-selection scheme that evaluates ensembles selected by several DES techniques using different metrics. Experimental results show that using accuracy as a metric to select the ensembles, PS-DES performs better than individual DES techniques. PS-DES source code is available in a GitHub repository$^{4}$.
\end{abstract}
%%%%%%%%%%%%%%%%%%%%%%%%%%%%%%%%%%%%%%%%%%%%%%%%%%%%%%%%%%%%%%%%%%%%%%%%%%%%%%%%

\section{Introduction}

Multiple Classifier Systems (MCS) are often used to improve the accuracy and reliability of machine learning models~\cite{cruz2018}. MCS has three main phases: generation, selection, and combination. In the generation phase, a pool of classifiers is created by training base classifiers with techniques such as Bagging~\cite{breiman1996bagging}, different models, and variations of learning algorithms~\cite{tsoumakas2005}.

The selection phase selects the most competent classifiers in the pool to predict a given query sample. Two main approaches for selecting classifiers are static selection~\cite{ko2008} and dynamic selection (DS)~\cite{cruz2018}. In the static approach, the selection is performed during training and works on the classifiers' overall performance on a validation set. In contrast, the DS approach selects the classifiers on the fly based on their competence in predicting a specific query sample. When only one classifier is selected, it is called Dynamic Classifier Selection~(DCS), and when more than one classifier is selected, it is called Dynamic Ensemble Selection~(DES). Examples of DES techniques include META-DES~\cite{cruz_metaDES2015}, Dynamic Ensemble Selection Performance (DES-P)~\cite{WOLOSZYNSKI2012207}, and K-Nearest Oracles Union (KNORA-U)~\cite{ko2008}. The last phase of an MCS is combination, also called integration. In this phase, the output of all the classifiers selected is combined to produce a final prediction. The combination or integration of the predictions can be done in various ways, including voting and weighting~\cite{zhou2012}.

Currently, research efforts in DES are focused on proposing new methods to improve phases, such as generation~\cite{MONTEIRO2023567}, selection~\cite{ELMI2021}, and combination~\cite{costa2018}. Additionally, there have been attempts to apply DES to other areas of knowledge~\cite{Swaminathan2022}. 
Despite the progress that has been made, no DES technique is suitable for all problems. This is in line with the statistical rationale for MCS~\cite{kuncheva2014}, which suggests that combining multiple classifiers increases the likelihood of finding the optimal result for any given problem. However, to the authors' knowledge, the field still lacks techniques that work on evaluating the ensembles selected by DES methods and explores the advantages of pre-selected ensembles to obtain better performance.

Aiming to evaluate this gap in DES' research field, we pose the following research question: ``How to analyze ensembles selected by different DES techniques and choose the one having the highest correct prediction potential?" To investigate this question, we propose the Post-Selection Dynamic Ensemble Selection~(PS-DES) approach. PS-DES is based on the assumption that different selection criteria may lead to different selected ensembles, and the best criteria used to select an ensemble may differ on an instance-to-instance basis. PS-DES aims to analyze and choose the best ensemble from a set of ensembles generated by various DES techniques to obtain more reliable predictions. Therefore, our proposal work as a post-selection scheme, i.e., it performs after the selection phase of different DES methods and before the combination phase. 

Moreover, the best ensemble is selected based on a new concept of ensemble potential proposed in this work. In contrast to the selection criteria employed in many DES methods such as META-DES~\cite{cruz_metaDES2015} that work by estimating the quality or competence of each model, the proposed ensemble potential evaluates whether the final selected ensemble of classifiers is reliable. We propose three approaches based on classical performance estimation metrics for measuring the ensemble potential: Accuracy, F-score, and Matthew's Correlation Coefficient.

Experiments over 20 classification datasets and considering three different performance evaluation metrics demonstrate that the post-selection scheme based on the ensemble potential leads to systematic improvement in classification performance over state-of-the-art DES methods. Thus, the evaluation of the pre-selected ensemble capabilities should not be neglected.
The rest of the paper is organized as follows: Section II shows a literature review on DES. Section III presents our proposal. Section IV shows the experimental setup. The results are discussed in Section V, and Section VI presents the conclusions.

\section{Literature review}
Typically, the development of a DES involves three stages. Firstly, in the generation phase, a pool of classifiers is generated. Secondly, in the selection phase, a subset of classifiers (ensemble) is chosen from the pool created in the generation phase. Finally, the classifiers in the ensemble are combined to classify a given query sample in the combination, or integration, phase.

During DES' generation phase, a pool of classifiers is created, denoted as $P = \{C_1, C_2, \ldots, C_M\}$, where $M$ represents the number of classifiers in the pool. The classifiers in the pool must exhibit both diversity and accuracy. Diversity \cite{kuncheva2014} refers to the property that the classifiers should not make the same prediction mistakes, as this is crucial to cover the feature space adequately. Several approaches can be used to generate a pool. These approaches include using different distributions of the training set, such as Bagging \cite{breiman1996bagging}, using different parameters for the same base classifier (e.g., variations in the number of neighbors in a k-Nearest Neighbors algorithm), or using different base classifiers altogether, which are called heterogeneous ensembles \cite{tsoumakas2005}. Heterogeneous ensembles tend to be more diverse than homogeneous ones due to their different mathematical formulations, which typically result in different classification results \cite{wang2021}.

The second phase of developing a DES is selection, which aims to choose a subset of classifiers ($ P' \subseteq P $), also known as an ensemble. There are two approaches to selection, namely static and dynamic \cite{cruz2018}. A fixed subset of classifiers is chosen for all test samples in the static approach. In contrast, the dynamic approach, called Dynamic Ensemble Selection (DES), involves selecting a subset of the pool for each query sample $\mathbf{x}_q$. In dynamic selection, classifiers are chosen based on some criteria, given the pool created in the previous phase. Among the criteria found in the literature are the Oracle approach, as seen in KNORA-E, KNORA-U \cite{ko2008}, and K-Nearest Output Profile (KNOP) \cite{cavalin2013dynamic}, accuracy-based methods, such as DES Performance (DES-P) \cite{WOLOSZYNSKI2012207}, and meta-learning, as in the case of META-DES \cite{cruz_metaDES2015}. These criteria are typically computed from the Region of Competence (RoC), a local region for a query sample ($\mathbf{x}_q$), denoted as $\theta_{\mathbf{x}_q}$, which is a fundamental concept in dynamic selection approaches. The RoC is usually obtained by applying k-NN or clustering methods to a validation set (DSEL) or the training set itself, such that $\theta_{\mathbf{x}_q} = \{\mathbf{x}_1, \ldots, \mathbf{x}_k\}$, where $k$ is the size of the ROC.

The final phase of a DES is integration, also called aggregation or combination, which involves combining the classifiers selected in the selection phase when multiple classifiers are chosen. Techniques used in this phase include majority vote, product rule, and sum rule~\cite{kuncheva2014}.

It is worth noting that research papers related to DES do not focus on assessing ensembles that DES techniques have already generated. Elmi and Eftekhari \cite{ELMI2021} presented a solution by utilizing the selection phase of DES approaches. However, their proposed approach only allowed for layer-by-layer ensemble selection and did not allow the evaluation of the collective output of all ensembles generated by DES methods.

\section{Post-Selection Dynamic Ensemble Selection}

The proposed Post-Selection Dynamic Ensemble Selection~(PS-DES) is based on the notion of potential, the capability of an ensemble selected by a given DS technique to make a correct prediction. Consequently, it works as a post-processing scheme for ensembles chosen according to different criteria (e.g., meta-learning, Oracle, accuracy). In addition, this proposal aims to evaluate the quality or potential of a selected ensemble, which contrasts with the current DES methods that build an ensemble by selecting multiple competent classifiers individually without trying to characterize the selected dynamic ensemble. This approach consists of three phases: \textbf{(1)} pool generation and setup, \textbf{(2)} post-selection, and \textbf{(3)} combination.

\subsection{Phase 1: Pool generation and DES' setup}
\label{sec:geracao_pool}

The initial stage of PS-DES involves generating a pool of classifiers and configuring the DES techniques. First, Bagging generates multiple bootstraps ($T^{b}$) from the original dataset ($T$), where $b$ is the number of bootstraps. Then, a pool of $b \times m$ classifiers, denoted as $P = \{{C_{1}^{1}, C_{1}^{2}, \ldots, C_{m}^{b}\}}$, is constructed by training each of the $m$ classifiers ($C_{1}, \ldots, C_{m}$) on each of the $b$ bootstraps generated by Bagging. Finally, any DES techniques specified by the user, including META-DES, KNORA-U, and DES-P, are initialized using the same pool $P$ as input. These techniques can be used to select an intermediate ensemble later validated by our post-processing scheme to obtain the optimal one. All DES approaches,  $DES_{set} = \{des_1, des_2, \ldots, des_n\}$, are consolidated into the set $DES_{set}$.

\subsection{Phase 2: Selection}

This phase seeks to identify the optimal ensemble from a set of ensembles generated by different DES techniques, and Algorithm~\ref{alg:ho_des_sel} shows its pseudo-code. Given a query sample ($\mathbf{x}_q$), the validation dataset $DSEL$, and a set of DES techniques ($DES_{set}$), this phase involves selecting several dynamic ensembles $P'$, each one created using a different DES method, and assessing their effectiveness in order to determine which ensemble is most likely to perform well for the given $\mathbf{x}_q$.

\begin{algorithm} 
\caption{PS-DES selection}
\label{alg:ho_des_sel} 
\begin{algorithmic}[1] 
\Procedure{PS-DES\_Selection}{$\mathbf{x_q},DSEL, DES_{set}$} 
\State $pot_{max} \gets 0 $
\State $P^{sel} \gets \emptyset $
\State $\theta_{\mathbf{x}_q} \gets calculate\_{ROC}(\mathbf{x}_q,DSEL)$

\For {$des$ in $DES_{set}$} 
    \State $P' \gets get\_ensemble(des, \theta_{\mathbf{x}_q})$ 
    \State $ pot_{des} \gets calculate\_pot(P')$
    \If{$pot_{des} \geq pot_{max}$} 
        \State $ pot_{max} \gets pot_{des}$
        \State $P^{sel} \gets P' $ 
    \EndIf 
\EndFor
\State \textbf{return} $P^{sel}$    
\EndProcedure 
\end{algorithmic} 
\end{algorithm}

This phase begins by computing the Region on Competence (RoC) ($\theta_{\mathbf{x}_q}$) for the query sample $\mathbf{x}_q$ using the k-NN algorithm. It is essential to note that all $DES_{n}$ utilize the same RoC ($\theta_{\mathbf{x}_q}$) based on the k-NN, thereby reducing the computational burden of implementing multiple selection criteria.

Then, for each technique in $DES_{set}$, a set of classifiers is selected according to its competence estimation and selection criterion, forming the ensemble $P'$ (lines 5 and 6). Subsequently, the potential of the generated ensemble $P'$ is evaluated (line 7). As the class label of $\mathbf{x}_q$ is unknown, the potential assumes that the output class of $\mathbf{x}_q$ corresponds to the majority vote of the ensemble. Consequently, it computes the potential of this ensemble by assessing the proportion of methods in it that contribute to this decision. For instance, given a binary classification problem and an ensemble with seven base classifiers, $P' = \{C_1, \ldots, C_7\}$ selected by a given DES technique, and let $y_{P'} = \{0,1,0,0,1,1,1\}$ be the predictions of the base classifier for the given $\mathbf{x}_q$. The majority vote would give the class $1$ as the answer. The potential is then estimated based on a classical performance metric comparing the ensemble majority vote and the votes of each classifier using a performance metric. If accuracy is used to calculate $P'$ potential, the value would be $pot_{des} = 0.57$. If the F-score is chosen as the potential metric, the $P'$ potential is $pot_{des} = 0.73$. 

After evaluating the potential of all possible ensembles, the one that obtained the highest value, $P^{sel}$, is returned as the selected one for the combination step.

\subsection{Phase 3: Combination}

Once the $P^{sel}$ selection is complete, Phase 3 begins, which is accountable for combining the classifiers into $P^{sel}$, using techniques such as majority vote or sum rule.

\section{Experimental setup}

%\subsection{Datasets}
\noindent \textbf{Datasets.}
The experiments were conducted using 20 datasets from the UCI Machine Learning Repository~\cite{uci}, which vary in sample size, dimensions, number of classes, and Imbalance Ratio (IR) (Table~\ref{tab:bases}). Each dataset $T$ is split into three parts: training (50\%), $DSEL$ (25\%), and testing (25\%). This split is stratified, meaning that the proportions of the classes between the three datasets are maintained. For each dataset, we run 30 replications, changing the distribution of the sets (holdout) to obtain the average values for the evaluated metrics. The data is scaled using the Standard Scaler (also known as Z-score normalization~\cite{de2023choice}).

\begin{table}[!htp]
\centering
\caption{Datasets main characteristics. The number of samples, dimensions (Dim), classes, and Imbalance Ratio (IR).}
\begin{tabular}{|l|l|l|l|l|}
\hline
\textbf{Datasets} & \textbf{Examples} & \textbf{Dim} & \textbf{Classes} & \textbf{IR} \\ \hline
appendicitis  & 106              & 7                  & 2                & 2.52        \\ \hline
australian    & 690              & 14                 & 2                & 1.12        \\ \hline
balance       & 625              & 4                  & 3                & 2.63        \\ \hline
cmc           & 1473             & 9                  & 3                & 1.30        \\ \hline
column\_3C     & 310              & 6                  & 3                & 2.50        \\ \hline
diabetes      & 768              & 8                  & 2                & 1.86        \\ \hline
glass1        & 214              & 9                  & 2                & 1.82        \\ \hline
glass6        & 214              & 9                  & 2                & 6.38        \\ \hline
haberman      & 306              & 3                  & 2                & 2.78        \\ \hline
hayes    & 160              & 4                  & 3                & 3.40        \\ \hline
heart         & 270              & 13                 & 2                & 1.25        \\ \hline
led7digit     & 500              & 7                  & 10               & 1.54        \\ \hline
mammographic  & 830              & 5                  & 2                & 1.15        \\ \hline
musk          & 476              & 166                & 2                & 1.29        \\ \hline
pima          & 768              & 8                  & 2                & 1.90        \\ \hline
sonar         & 208              & 60                 & 2                & 1.14        \\ \hline
vehicle       & 846              & 18                 & 4                & 1.10        \\ \hline
vehicle2      & 846              & 18                 & 2                & 2.88        \\ \hline
vowel         & 990              & 13                 & 11               & 1.00        \\ \hline
wdbc          & 683              & 9                  & 2                & 1.85        \\ \hline
\end{tabular}%
\label{tab:bases}
\end{table}

%\subsection{Phase 1}
\noindent \textbf{Phase 1.}
First, Bagging, 100 bootstraps were used for all experiments, consistent with previous studies \cite{cruz2018, ko2008, cruz_metaDES2015}. For the pool generation, three base classifiers (Perceptron, Logistic Regression, and Naive Bayes) were selected for the experiments. As they have different mathematical foundations and low computational costs \cite{Nguyen2020,costa2018, cruz2018} they are suitable for building a diverse and lightweight pool of classifiers. Thus, the classifier pool ($P$) consisted of 300 classifiers ($\text{3 base classifiers} \times \text{100 bootstraps}$). Since the focus of the research was not on optimizing each base model's hyperparameters, the default hyperparameters values from scikit-learn were used. 
 
Four DES approaches (KNORA-U, KNOP, DES-P, and META-DES) were chosen due to their application of various selection criteria (e.g., Oracle, accuracy, meta-learning). These approaches showed superior performance in a recent empirical study \cite{cruz2018}. We applied these DES methods default hyperparameter configurations of the DESlib 0.3 library~\cite{deslib} to guarantee experiment consistency. Moreover, the same pool of classifiers was utilized to fairly compare all DES techniques.

\noindent \textbf{Phase 2.}
The Region of Competence (RoC) was calculated applying k-Nearest Neighbors (k-NN) with k = 7, as suggested in \cite{cruz2018}. To assess the performance of the ensembles, we employed a range of evaluation metrics, including accuracy, F-score, and Matthews Correlation Coefficient (MCC). Accuracy is a popular metric for DES techniques, although it may not be suitable for imbalanced datasets (i.e., high IR). Meanwhile, F-score and MCC are more suitable for such datasets. F-score is advantageous in scenarios where there is an appreciation for recall and precision, since these two metrics are used in its calculation. The MCC considers false-negative rates in its formulation, which can be of interest to specific problems.

The PS-DES variants are labeled according to the metric used to calculate the potential: accuracy (PS-DES-acc), F-score (PS-DES-F), and Matthews Correlation Coefficient (PS-DES-MCC). To assess whether the proposed metrics perform better than random selection, we also conducted an experiment that randomly selected the best ensemble (PS-DES-Random).

\noindent \textbf{Phase 3.}
%\subsection{Phase 3}
Finally, majority voting was used as a combination approach since individual DES techniques usually apply it~\cite{cruz2018}.
  
\section{Results and discussions}

The proposed method is evaluated based on three metrics: accuracy (Table~\ref{tab:acc}), F-score (Table~\ref{tab:f-score}), and MCC (Table~\ref{tab:mcc}). Upon examining the tables, our results indicate that PS-DES-acc outperforms all the other approaches in all metrics. The PS-DES-acc obtained the best rank considering all performance metrics, followed by the variant using the F-score metric for computing the ensemble's potential. These results are interesting since, even though the final proposal may be evaluated regarding a different performance metric (e.g., F-score or MCC), using accuracy as the metric for computing the ensemble potential is more advantageous. 

Analogously, MCC obtained the lowest ranking among all PS-DES variants even when in the scenario that MCC is used as a performance evaluation metric to compute the overall method performance. This result indicates no relation between the metrics selected for calculating ensemble potential and the same metric applied to evaluate the approaches. For accuracy, F-score, and MCC, the chosen metric for calculating the potential does not interfere with the approach's evaluation. Nevertheless, according to these tables, the average ranking of PS-DES approaches is systematically better when compared to individual DS techniques (e.g., META-DES). Thus, the proposed post-processing selection scheme indeed leads to more robust dynamic ensemble selection systems.

%regarding ranking, followed by PS-DES-F

However, to see if such a difference in performance is significant, we need to go further and perform a more fine-grained analysis by comparing pair of techniques over multiple datasets. Hence, we also conducted one analysis based on the number of wins, ties, and losses (w/t/l) obtained by a control technique and the Wilcoxon signed rank test with a confidence level of 95\%. Results of these pairwise comparisons are presented in Tables~\ref{tab:stat_acc},~\ref{tab:stat_f}, and~\ref{tab:stat_mcc} for the PS-DES-acc, PS-DES-F and PS-DES-MCC methods, respectively.

%RMOC - Mudei as tabelas para outro arquivo para facilitar a escrita.
\begin{table*}[!htp]
\centering
%\caption{Average accuracy evaluation for DES techniques: KNORA-U (KNRU), KNOP, META-DES (M-DES), DES-P, PS-DES-Random (Random), PS-DES-MCC, PS-DES-F, and PS-DES-acc. The best result from each dataset is presented in bold.}
\caption{Average accuracy and rankings of the evaluated methods for each dataset. The best result per dataset is presented in bold.}
\begin{tabular}{lrrrrrrrr}
\hline
\textbf{Datasets} & \multicolumn{1}{l}{\textbf{KNORA-U}} & \multicolumn{1}{l}{\textbf{KNOP}} & \multicolumn{1}{l}{\textbf{META-DES}} & \multicolumn{1}{l}{\textbf{DES-P}} & \multicolumn{1}{l}{\textbf{Random}} & \multicolumn{1}{l}{\textbf{PS-DES-MCC}} & \multicolumn{1}{l}{\textbf{PS-DES-F}} & \multicolumn{1}{l}{\textbf{PS-DES-acc}} \\ \hline
appendicitis   & 0.859                              & 0.862                             & 0.851                              & 0.864                              & 0.855                               & 0.859                                 & 0.868                               & \textbf{0.869}                        \\
australian     & 0.847                              & 0.846                             & 0.848                              & 0.846                              & 0.846                               & \textbf{0.849}                        & \textbf{0.849}                      & \textbf{0.849}                        \\
balance        & 0.883                              & 0.886                             & \textbf{0.893}                     & 0.883                              & 0.885                               & \textbf{0.893}                        & 0.891                               & 0.889                                 \\
cmc            & 0.509                              & 0.512                             & 0.488                              & \textbf{0.513}                     & 0.506                               & 0.483                                 & 0.505                               & 0.510                                 \\
column\_3C     & 0.845                              & 0.844                             & 0.841                              & \textbf{0.848}                     & 0.846                               & 0.839                                 & 0.845                               & 0.847                                 \\
diabetes       & 0.768                              & 0.769                             & 0.762                              & 0.765                              & 0.767                               & 0.762                                 & 0.767                               & \textbf{0.770}                        \\
glass1         & 0.642                              & 0.640                             & 0.674                              & \textbf{0.682}                     & 0.653                               & \textbf{0.682}                        & 0.672                               & 0.672                                 \\
glass6         & 0.940                              & 0.938                             & 0.947                              & 0.944                              & 0.943                               & \textbf{0.948}                        & \textbf{0.948}                      & 0.947                                 \\
haberman       & 0.731                              & 0.731                             & 0.726                              & \textbf{0.737}                     & 0.734                               & 0.727                                 & \textbf{0.737}                      & 0.734                                 \\
hayes          & 0.611                              & 0.619                             & 0.670                              & 0.640                              & 0.640                               & 0.665                                 & \textbf{0.684}                      & 0.682                                 \\
heart          & 0.836                              & 0.838                             & 0.833                              & \textbf{0.841}                     & 0.836                               & 0.837                                 & 0.838                               & 0.837                                 \\
led7digit      & 0.725                              & \textbf{0.726}                    & 0.708                              & 0.723                              & 0.721                               & 0.697                                 & 0.704                               & 0.723                                 \\
mammographic   & 0.830                              & \textbf{0.831}                    & 0.822                              & 0.828                              & 0.830                               & 0.827                                 & 0.824                               & 0.825                                 \\
musk           & 0.779                              & 0.781                             & 0.788                              & 0.792                              & 0.786                               & 0.790                                 & 0.791                               & \textbf{0.794}                        \\
pima           & \textbf{0.770}                     & 0.768                             & 0.761                              & 0.766                              & 0.766                               & 0.762                                 & 0.763                               & 0.766                                 \\
sonar          & 0.771                              & 0.771                             & 0.796                              & 0.784                              & 0.787                               & 0.787                                 & 0.796                               & \textbf{0.798}                        \\
vehicle        & 0.750                              & 0.752                             & \textbf{0.761}                     & 0.752                              & 0.750                               & 0.755                                 & 0.758                               & \textbf{0.761}                        \\
vehicle2       & 0.949                              & 0.947                             & \textbf{0.956}                     & 0.951                              & 0.951                               & 0.955                                 & 0.954                               & 0.955                                 \\
vowel          & 0.965                              & 0.965                             & \textbf{0.981}                     & 0.967                              & 0.969                               & 0.980                                 & 0.978                               & 0.977                                 \\
wdbc           & \textbf{0.970}                     & \textbf{0.970}                    & 0.969                              & \textbf{0.970}                     & \textbf{0.970}                      & 0.968                                 & 0.969                               & 0.968                                 \\ \hline
ranking        & 5.45                      & 5.03                              & 4.90                               & 3.88                               & 5.23                                & 4.73                                  & 3.70                                & 3.10                                  \\
\hline
\end{tabular}%
\label{tab:acc}
\end{table*}

\begin{table*}[!htp]
\centering
%\caption{Average F-score evaluation for DES techniques: KNORA-U (KNRU), KNOP, META-DES (M-DES), DES-P, PS-DES-Random (Random), PS-DES-MCC, PS-DES-F, and PS-DES-acc. The best result from each dataset is presented in bold.}
\caption{Average F-score and rankings of the evaluated methods for each dataset. The best result per dataset is presented in bold.}
\begin{tabular}{lrrrrrrrr}
\hline
\textbf{Datasets} & \multicolumn{1}{l}{\textbf{KNORA-U}} & \multicolumn{1}{l}{\textbf{KNOP}} & \multicolumn{1}{l}{\textbf{META-DES}} & \multicolumn{1}{l}{\textbf{DES-P}} & \multicolumn{1}{l}{\textbf{Random}} & \multicolumn{1}{l}{\textbf{PS-DES-MCC}} & \multicolumn{1}{l}{\textbf{PS-DES-F}} & \multicolumn{1}{l}{\textbf{PS-DES-acc}} \\ \hline
appendicitis     & 0.708                              & 0.718                             & 0.717                              & 0.749                              & 0.715                               & 0.732                                 & \textbf{0.760}                      & 0.754                                 \\
australian       & 0.845                              & 0.843                             & 0.845                              & 0.844                              & 0.843                               & 0.846                                 & \textbf{0.847}                      & 0.846                                 \\
balance          & 0.612                              & 0.615                             & 0.631                              & 0.613                              & 0.615                               & \textbf{0.637}                        & 0.628                               & 0.626                                 \\
cmc              & 0.472                              & 0.477                             & 0.454                              & \textbf{0.484}                     & 0.472                               & 0.451                                 & 0.474                               & 0.478                                 \\
column\_3C        & 0.799                              & 0.796                             & 0.792                              & \textbf{0.803}                     & 0.800                               & 0.790                                 & 0.798                               & 0.801                                 \\
diabetes         & 0.725                              & 0.725                             & 0.719                              & 0.723                              & 0.724                               & 0.719                                 & 0.726                               & \textbf{0.729}                        \\
glass1           & 0.422                              & 0.415                             & \textbf{0.585}                     & 0.577                              & 0.512                               & 0.584                                 & 0.530                               & 0.535                                 \\
glass6           & 0.850                              & 0.847                             & \textbf{0.880}                     & 0.868                              & 0.863                               & 0.879                                 & 0.878                               & 0.873                                 \\
haberman         & 0.506                              & 0.507                             & 0.525                              & 0.542                              & 0.519                               & 0.520                                 & \textbf{0.555}                      & 0.549                                 \\
hayes            & 0.627                              & 0.634                             & 0.688                              & 0.658                              & 0.654                               & 0.681                                 & 0.697                               & \textbf{0.699}                        \\
heart            & 0.832                              & 0.834                             & 0.829                              & \textbf{0.837}                     & 0.832                               & 0.833                                 & 0.834                               & 0.833                                 \\
led7digit        & 0.718                              & \textbf{0.719}                    & 0.700                              & 0.715                              & 0.714                               & 0.687                                 & 0.693                               & \textbf{0.719}                        \\
mammographic     & 0.829                              & \textbf{0.830}                    & 0.821                              & 0.826                              & 0.829                               & 0.825                                 & 0.822                               & 0.824                                 \\
musk             & 0.770                              & 0.772                             & 0.781                              & 0.785                              & 0.778                               & 0.783                                 & 0.784                               & \textbf{0.787}                        \\
pima             & \textbf{0.727}                     & 0.724                             & 0.718                              & 0.724                              & 0.722                               & 0.718                                 & 0.721                               & 0.725                                 \\
sonar            & 0.768                              & 0.767                             & 0.792                              & 0.780                              & 0.783                               & 0.784                                 & 0.793                               & \textbf{0.795}                        \\
vehicle          & 0.742                              & 0.745                             & \textbf{0.757}                     & 0.744                              & 0.743                               & 0.751                                 & 0.753                               & 0.755                                 \\
vehicle2         & 0.933                              & 0.931                             & \textbf{0.943}                     & 0.936                              & 0.935                               & 0.941                                 & 0.939                               & 0.941                                 \\
vowel            & 0.886                              & 0.886                             & \textbf{0.940}                     & 0.894                              & 0.899                               & 0.936                                 & 0.929                               & 0.929                                 \\
wdbc             & 0.966                              & 0.966                             & 0.965                              & \textbf{0.967}                     & \textbf{0.967}                      & 0.965                                 & 0.966                               & 0.964                                 \\ \hline
ranking          & 5.80                               & 5.45                              & 4.45                               & 3.95                               & 5.45                                & 4.60                                  & 3.45                                & 2.85                                  \\ \hline
\end{tabular}%
\label{tab:f-score}
\end{table*}

\begin{table*}[!ht]
\centering
%\caption{Average MCC evaluation for DES techniques: KNORA-U (KNRU), KNOP, META-DES (M-DES), DES-P, PS-DES-Random (Random), PS-DES-MCC, PS-DES-F, and PS-DES-acc. The best result from each dataset is presented in bold.}
\caption{Average MCC and rankings of the evaluated methods for each dataset. The best result per dataset is presented in bold.}
\begin{tabular}{lrrrrrrrr}
\hline
\textbf{Datasets} & \multicolumn{1}{l}{\textbf{KNORA-U}} & \multicolumn{1}{l}{\textbf{KNOP}} & \multicolumn{1}{l}{\textbf{META-DES}} & \multicolumn{1}{l}{\textbf{DES-P}} & \multicolumn{1}{l}{\textbf{Random}} & \multicolumn{1}{l}{\textbf{PS-DES-MCC}} & \multicolumn{1}{l}{\textbf{PS-DES-F}} & \multicolumn{1}{l}{\textbf{PS-DES-acc}} \\ \hline
appendicitis     & 0.503                              & 0.517                             & 0.489                              & 0.563                              & 0.500                               & 0.529                                 & \textbf{0.574}                      & 0.566                                 \\
australian       & 0.694                              & 0.691                             & 0.694                              & 0.691                              & 0.690                               & 0.695                                 & \textbf{0.697}                      & 0.695                                 \\
balance          & 0.794                              & 0.801                             & 0.812                              & 0.795                              & 0.798                               & \textbf{0.813}                        & 0.809                               & 0.806                                 \\
cmc              & 0.227                              & 0.232                             & 0.196                              & \textbf{0.238}                     & 0.224                               & 0.187                                 & 0.223                               & 0.232                                 \\
column\_3C        & 0.755                              & 0.753                             & 0.748                              & \textbf{0.759}                     & 0.756                               & 0.744                                 & 0.755                               & 0.757                                 \\
diabetes         & 0.466                              & 0.468                             & 0.454                              & 0.460                              & 0.464                               & 0.453                                 & 0.466                               & \textbf{0.472}                        \\
glass1           & 0.019                              & 0.008                             & 0.226                              & 0.242                              & 0.134                               & \textbf{0.247}                        & 0.189                               & 0.196                                 \\
glass6           & 0.729                              & 0.723                             & 0.770                              & 0.755                              & 0.749                               & \textbf{0.775}                        & 0.774                               & 0.764                                 \\
haberman         & 0.131                              & 0.130                             & 0.144                              & 0.193                              & 0.157                               & 0.138                                 & \textbf{0.201}                      & 0.185                                 \\
hayes            & 0.416                              & 0.427                             & 0.505                              & 0.467                              & 0.461                               & 0.499                                 & \textbf{0.529}                      & 0.524                                 \\
heart            & 0.673                              & 0.677                             & 0.667                              & \textbf{0.682}                     & 0.673                               & 0.674                                 & 0.677                               & 0.674                                 \\
led7digit        & 0.696                              & \textbf{0.697}                    & 0.678                              & 0.693                              & 0.692                               & 0.666                                 & 0.674                               & 0.694                                 \\
mammographic     & 0.660                              & \textbf{0.662}                    & 0.644                              & 0.655                              & 0.660                               & 0.652                                 & 0.647                               & 0.650                                 \\
musk             & 0.547                              & 0.551                             & 0.566                              & 0.575                              & 0.562                               & 0.572                                 & 0.574                               & \textbf{0.579}                        \\
pima             & \textbf{0.471}                     & 0.466                             & 0.451                              & 0.462                              & 0.461                               & 0.452                                 & 0.455                               & 0.464                                 \\
sonar            & 0.547                              & 0.544                             & 0.594                              & 0.570                              & 0.576                               & 0.575                                 & 0.595                               & \textbf{0.600}                        \\
vehicle          & 0.672                              & 0.673                             & \textbf{0.684}                     & 0.674                              & 0.671                               & 0.676                                 & 0.681                               & \textbf{0.684}                        \\
vehicle2         & 0.867                              & 0.863                             & \textbf{0.886}                     & 0.873                              & 0.872                               & 0.884                                 & 0.880                               & 0.883                                 \\
vowel            & 0.784                              & 0.782                             & \textbf{0.885}                     & 0.798                              & 0.808                               & 0.877                                 & 0.864                               & 0.864                                 \\
wdbc             & 0.933                              & 0.933                             & 0.931                              & \textbf{0.934}                     & 0.933                               & 0.931                                 & 0.932                               & 0.929                                 \\ \hline
ranking          & 5.45                               & 5.30                              & 4.80                               & 3.80                               & 5.35                                & 4.65                                  & 3.55                                & 3.10                                  \\
\hline
\end{tabular}%
\label{tab:mcc}
\end{table*}

\begin{table*}[!htp]
\centering
\caption{Statistical Analyses for PS-DES-Acc against state-of-the-art DS methods. The line (w/t/l) presents the number of wins, ties, and losses it obtained compared to the column-wise technique. The p-value line shows the result of applying the paired Wilcoxon statistical test. Statistically different results ($\alpha=0.5$) are highlighted in bold.}
\begin{tabular}{llrrrrrrr}
\hline
\textbf{Metric} &         & \multicolumn{1}{l}{\textbf{KNORA-U}} & \multicolumn{1}{l}{\textbf{KNOP}} & \multicolumn{1}{l}{\textbf{META-DES}} & \multicolumn{1}{l}{\textbf{DES-P}} & \multicolumn{1}{l}{\textbf{Random}} & \multicolumn{1}{l}{\textbf{PS-DES-MCC}} & \multicolumn{1}{l}{\textbf{PS-DES-F}} \\ \hline
\multirow{2}{*}{Accuracy} & w/t/l   & 16/0/4                               & 14/0/6                            & 13/0/7                                & 13/0/7                             & 17/1/2                              & 12/1/7                                & 12/0/8                              \\
                          & p-value & \textbf{0.002}                                & \textbf{0.003}                             & \textbf{0.005}                                 & 0.062                              & \textbf{0.001}                               & \textbf{0.016}                                 & 0.071                               \\
\multirow{2}{*}{F-score}  & w/t/l   & 17/0/3                               & 17/0/3                            & 13/0/7                                & 14/0/6                             & 18/0/2                              & 12/0/8                                & 12/0/8                              \\
                          & p-value & \textbf{0.000}                                & \textbf{0.000}                             & \textbf{0.049}                                 & \textbf{0.020}                              & \textbf{0.000}                               & 0.066                                 & 0.139                               \\
\multirow{2}{*}{MCC}      & w/t/l   & 16/0/4                               & 14/0/6                            & 13/0/7                                & 13/0/7                             & 18/0/2                              & 13/0/7                                & 11/0/9                              \\
                          & p-value & \textbf{0.001}                                & \textbf{0.001}                             & \textbf{0.029}                                 & 0.077                              & \textbf{0.000}                              & \textbf{0.049}                                & 0.261                               \\ \hline
\end{tabular}%
\label{tab:stat_acc}
\end{table*}

\begin{table*}[!htp]
\centering
\caption{Statistical Analyses for PS-DES-F against state-of-the-art DS methods. The line (w/t/l) presents the number of wins, ties, and losses that it obtained compared to the column-wise technique. The p-value line shows the result of applying the paired Wilcoxon statistical test. Statistically different results ($\alpha=0.5$) are highlighted in bold.}
\begin{tabular}{llrrrrrrr}
\hline
\textbf{Metric} &         & \multicolumn{1}{l}{\textbf{KNORA-U}} & \multicolumn{1}{l}{\textbf{KNOP}} & \multicolumn{1}{l}{\textbf{META-DES}} & \multicolumn{1}{l}{\textbf{DES-P}} & \multicolumn{1}{l}{\textbf{Random}} & \multicolumn{1}{l}{\textbf{PS-DES-MCC}} & \multicolumn{1}{l}{\textbf{PS-DES-acc}} \\ \hline
\multirow{2}{*}{Accuracy} & w/t/l   & 13/1/6                               & 13/1/6                            & 13/0/7                                & 10/0/10                            & 13/0/7                              & 15/0/5                                & 8/0/12                                \\
                          & p-value & \textbf{0.015}                                &\textbf{0.032}                             & \textbf{0.049}                                 & 0.237                              & \textbf{0.018}                               & \textbf{0.024}                                 & 0.934                                 \\
\multirow{2}{*}{F-score}  & w/t/l   & 15/0/5                               & 15/0/5                            & 13/0/7                                & 11/0/9                             & 15/0/5                              & 14/0/6                                & 8/0/12                                \\
                          & p-value & \textbf{0.003}                              & \textbf{0.005}                             & 0.174                                 & 0.147                              & \textbf{0.003}                             & 0.066                                 & 0.869                                 \\
\multirow{2}{*}{MCC}      & w/t/l   & 13/0/7                               & 14/0/6                            & 14/0/6                                & 11/0/9                             & 14/0/6                              & 14/0/6                                & 9/0/11                                \\
                          & p-value & \textbf{0.010}                                & \textbf{0.013}                             & \textbf{0.045}                                 & 0.139                              & \textbf{0.005}                               & \textbf{0.041}                              & 0.751                                 \\ \hline
\end{tabular}%
\label{tab:stat_f}
\end{table*}

\begin{table*}[!htp]
\centering
\caption{Statistical Analyses for PS-DES-MCC against state-of-the-art DS methods. The line (w/t/l) presents the number of wins, ties, and losses that it obtained compared to the column-wise technique. The p-value line shows the result of applying the paired Wilcoxon statistical test. Statistically different results ($\alpha=0.5$) are highlighted in bold.}
\begin{tabular}{llrrrrrrr}
\hline
\textbf{Metric} &         & \multicolumn{1}{l}{\textbf{KNORA-U}} & \multicolumn{1}{l}{\textbf{KNOP}} & \multicolumn{1}{l}{\textbf{META-DES}} & \multicolumn{1}{l}{\textbf{DES-P}} & \multicolumn{1}{l}{\textbf{Random}} & \multicolumn{1}{l}{\textbf{PS-DES-F}} & \multicolumn{1}{l}{\textbf{PS-DES-acc}} \\ \hline
\multirow{2}{*}{Accuracy} & w/t/l    & 12/0/8                               & 10/0/10                           & 10/0/10                               & 9/0/11                             & 12/0/8                              & 5/0/15                              & 7/1/12                                \\
                          & p-value & 0.174                                & 0.205                             & 0.580                                 & 0.649                              & 0.273                               & 0.978                               & 0.984                                 \\
\multirow{2}{*}{F-score}  & w/t/l   & 13/0/7                               & 12/0/8                            & 7/0/13                                & 9/0/11                             & 13/0/7                              & 6/0/14                              & 8/0/12                                \\
                          & p-value & \textbf{0.032}                                & \textbf{0.045}                             & 0.861                                 & 0.522                              & 0.101                               & 0.938                               & 0.938                                 \\
\multirow{2}{*}{MCC}      & w/t/l   & 13/0/7                               & 12/0/8                            & 9/0/11                                & 9/0/11                             & 11/0/9                              & 6/0/14                              & 7/0/13                                \\
                          & p-value & 0.071                                & 0.088                             & 0.676                                 & 0.663                              & 0.194                               & 0.962                               & 0.955                                 \\ \hline
\end{tabular}%
\label{tab:stat_mcc}
\end{table*}
The pairwise statistical analysis of PS-DES-acc shows it outperforms KNORA-U, KNOP, META-DES, Random, and PS-DES-MCC regarding accuracy (Table \ref{tab:stat_acc}). No significant difference is observed between PS-DES-acc and DES-P or PS-DES-F. However, considering the presence of datasets with $IR > 1$, it is necessary to consider F-score and MCC. The F-score analysis reveals that PS-DES-acc outperforms all DES individual techniques and Random, with no significant difference to PS-DES-F and PS-DES-MCC. For MCC, PS-DES-acc performs exceptionally well and obtains significantly better results compared to all techniques apart from DES-P and PS-DES-F. Ultimately, this variant based on accuracy for computing the ensemble potential obtained more victories than all other models, regardless of the performance metric used.

The statistical analysis of PS-DES-F (Table \ref{tab:stat_f}) indicates that it performs better than KNORA-U, KNOP, and Random on all three metrics. However, no statistical difference is found for MCC when compared with DES-P. However, the win-tie-loss analysis demonstrates that the PS-DES-F systematically obtained more wins against the state-of-the-art DES techniques and the random selection scheme (between 13 to 15 wins over the 20 datasets). In contrast, the analysis of PS-DES-MCC (Table \ref{tab:stat_mcc}) presents the worst results compared to PS-DES-acc and PS-DES-F. Based on Wilcoxon's test analysis, PS-DES-MCC scores better than KNORA-U and KNOP only in F-score. The hypothesis that PS-DES-MCC scores better cannot be refuted for all other metrics and comparisons.

In summary, the results indicate that PS-DES-acc and PS-DES-F yield comparable outcomes. Still, PS-DES-acc holds a slight advantage over its competitor, particularly when it is compared against the state-of-the-art DES methods.

\section{Conclusion}

This work proposed a new Dynamic Ensemble Selection (DES) method: Post-Selection Dynamic Ensemble Selection (PS-DES). This method is based on the idea that the optimal criteria for ensemble selection may differ at the instance level leading to ensembles with different qualities or ``potentials". To this end, the approach evaluates the potential of ensembles chosen by various DES techniques to determine which is more suitable for labeling a given instance.

Experiments demonstrate no direct correlation between the metrics applied for calculating the ensemble potential and for evaluating the approaches, as the PS-DES-acc was found to achieve the best overall results in all cases. Additionally, PS-DES was consistently superior to the existing state-of-the-art DES techniques, which implies that evaluating the selected ensembles as a collective is more important than assessing and choosing each base classifier separately, as is the trend in most DES methods. Thus, post-processing approaches in DES are vital, and future works will explore new metrics for measuring the ensemble's potential.

\vspace{-0.7em}

\section*{Acknowledgment}

The authors would like to thank the Instituto Federal de Pernambuco, Coordenação de Aperfeiçoamento de Pessoal de Nível Superior (CAPES), Fundação de Amparo à Ciência e Tecnologia de Pernambuco (FACEPE), and the Natural Sciences and Engineering Research Council of Canada (NSERC).

\bibliographystyle{IEEEtran}
\bibliography{root}

\end{document}